\pdfoutput=1

\documentclass[11pt]{article}

\usepackage[final]{acl}

\usepackage{times}
\usepackage{latexsym}
\usepackage{amsmath} 

\usepackage{setspace}
\usepackage{amsmath}
\usepackage{amssymb}

\usepackage{mathtools}    
\usepackage{enumitem}     
\usepackage{setspace}     

\usepackage{array}
\usepackage{colortbl}
\usepackage{xcolor}

\usepackage[T1]{fontenc}

\usepackage[utf8]{inputenc}

\usepackage{microtype}

\usepackage{inconsolata}

\usepackage{graphicx}

\usepackage{graphicx}
\usepackage{amsmath}
\usepackage{times}
\usepackage{framed}
\usepackage{ragged2e}
\usepackage{bm}
\usepackage{soul}
\usepackage{latexsym}
\usepackage{subfigure}
\usepackage{amsthm}
\usepackage{graphicx}
\usepackage{float}
\usepackage{longtable}
\usepackage{booktabs} 
\usepackage{multirow}
\usepackage{multicol}
\usepackage{amssymb}
\usepackage{dashbox}%
\usepackage{multirow}
\usepackage{textcomp}
\usepackage{tcolorbox}
\usepackage{bbding}
\usepackage{bbm}
\usepackage{mathtools}
\usepackage{adjustbox}
\usepackage[ruled,vlined]{algorithm2e}
\usepackage{color, soul}
\usepackage{pifont}
\usepackage{array}
\newcolumntype{P}[1]{>{\centering\arraybackslash}p{#1}}
\newcolumntype{M}[1]{>{\centering\arraybackslash}m{#1}}
\usepackage{cleveref}
\crefname{section}{§}{§§}
\Crefname{section}{§}{§§}
\crefname{figure}{Figure}{Figure}
\Crefname{figure}{Figure}{Figure}
\crefname{table}{Table}{Table}
\Crefname{table}{Table}{Table}
\usepackage{tabularx}

\newcommand\ourdataset{\textsc{IFBench}\xspace}
\SetKwInput{Input}{Input}
\SetKwInput{Output}{Output}

\SetKwSty{textbf}

\title{Structural Reward Model: Enhancing Interpretability, Efficiency, and Scalability in Reward Modeling}

\author{\textbf{Xiaoyu Liu\textsuperscript{1,*}}, \textbf{Di Liang\textsuperscript{1,2,†}}, \textbf{Chang Dai\textsuperscript{†}}, \textbf{Hongyu Shan\textsuperscript{†}}, \textbf{Peiyang Liu\textsuperscript{‡}}, \textbf{Yonghao Liu\textsuperscript{†‡}} \\ \textbf{Muling Wu\textsuperscript{†}}, \textbf{Yuntao Li\textsuperscript{†}}, \textbf{Xianjie Wu\textsuperscript{‡‡}}, \textbf{LI Miao\textsuperscript{†}}, \textbf{Jiangrong Shen\textsuperscript{†‡‡}}, \textbf{Minlong Peng\textsuperscript{‡‡‡,2}}\\
  \textsuperscript{*}Northeastern University, Boston, \textsuperscript{†}Independent Developer, \textsuperscript{‡}Peiking University, \\ \textsuperscript{†‡}Jilin University, \textsuperscript{‡‡}Beihang University, \textsuperscript{†‡‡}Xi'an Jiaotong University,  \textsuperscript{‡‡‡}Baidu Inc\\
  \texttt{\{liu.xiaoyu7\}@northeastern.edu, \{liangd17,mlpeng16\}@fudan.edu.cn} \\}

\begin{document}
\maketitle
\footnotetext[1]{Equal contribution}
\footnotetext[2]{Corresponding author: Di Liang, Minlong Peng.}
\begin{abstract}
Reward Models (RMs) are key components for evaluating and guiding language model outputs. However, traditional scalar RMs often struggle with incorporating contextual and background information during inference, leading to incomplete evaluations. Generative RMs (GRMs) attempt to address these limitations by generating intermediate reasoning steps. Yet, their uncontrolled black-box nature and inefficiency due to sequential decoding hinder their industrial deployment. Industrial scenarios, such as search and recommendation systems, often involve single-domain tasks requiring evaluation along specific dimensions. In such contexts, diagnosing “bad cases” necessitates structured feedback to identify and optimize dimension-specific issues.
In this paper, we propose the \textbf{Structural Reward Model (SRM)}, a modular and interpretable framework integrating side-branch models as auxiliary feature generators. By introducing fine-grained dimensions, SRMs enable interpretable and efficient evaluation, facilitating targeted diagnostics and optimization. This structured approach ensures adaptability and scalability for industrial applications.
Through comprehensive experiments, we demonstrate that SRMs outperform scalar RMs and GRMs in robustness and alignment with human preferences. The modular design further supports efficient optimization for practical scenarios, allowing SRM to provide a practical reward modeling solution for industry.

\end{abstract}

\section{Introduction}

Large Language Models (LLMs) have demonstrated remarkable capabilities in generating human-like text across diverse tasks~\cite{OpenAI2024}. However, ensuring these models deliver high-quality, contextually appropriate, and aligned responses continues to pose challenges~\cite{ouyang2022training, casper2023open}. Reward Models (RMs) play a vital role in addressing this by evaluating and guiding outputs based on human preferences~\cite{stiennon2020learning}. Traditional scalar RMs score responses using only the prompt and the generated output as input signals. While effective in many scenarios, this reliance on limited input often results in incomplete evaluations, as they lack access to richer contextual information and background knowledge during inference.

\begin{figure}[t]
\centering
\includegraphics[width=0.48\textwidth]{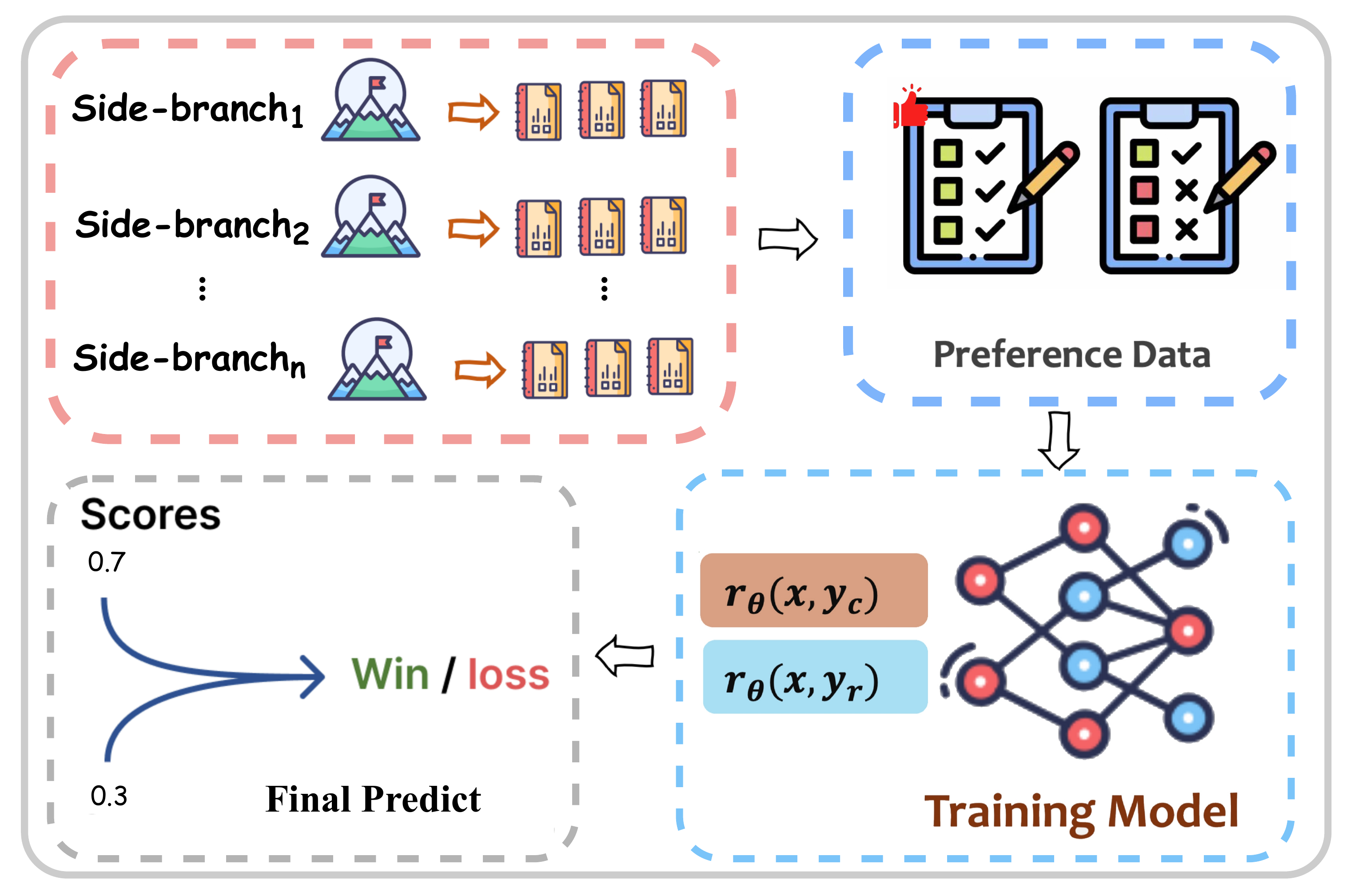}
\caption{Overview of the Side Branch Models Enhanced Structural Reward Model architecture. 
}
\label{fig:example}
\end{figure}

Recent advancements, such as Generative Reward Models (GRMs)~\cite{liu2025inferencetimescalinggeneralistreward, zhang2025generativeverifiersrewardmodeling,wu2025contrastive,wu2025llm}, attempt to mitigate the limitations of scalar RMs by generating intermediate reasoning steps to inform the reward evaluation process. Despite their conceptual promise, GRMs face substantial challenges in practical industrial deployment due to their uncontrolled, black-box nature and inefficiency stemming from sequential decoding~\cite{sinha2024challenges}. These characteristics hinder interpretability and scalability, reducing their applicability in real-world scenarios like search or recommendation systems.
For example, tasks in these industries often require assessments based on dimensions like relevance, timeliness, and authority~\cite{zhang2017building,wuimgfu}. Diagnosing “bad cases” in these settings necessitates structured feedback that can pinpoint specific dimensions for optimization~\cite{lee2015compass,wu2025image}. Without such interpretability, GRMs struggle to provide actionable insights.

To address the gaps in current reward modeling approaches, we introduce the \textbf{Structural Reward Model (SRM)} framework, as illustrated in Figure~\ref{fig:example}. The SRM integrates side-branch models as auxiliary feature generators inspired by the feature engineering principles from traditional machine learning paradigms~\cite{rewardbench}. Unlike scalar RMs that rely solely on prompt-response pairs or GRMs that operate as black-box generators, SRMs utilize modular, interpretable components capable of extracting fine-grained signals from input data. These side-branch models capture additional dimensions of contextual cues, such as semantic understanding, entity augmentation, style consistency, alignment with external knowledge, and diversity of responses~\cite{yao2025valuecompassbenchmarksplatform}. By combining these complementary features, the SRM transforms reward evaluation from a simple scalar rating to a more flexible rating process.

The structured nature of SRMs addresses the inefficiencies and interpretability limitations of GRMs by enabling feature-specific diagnostics. For instance, in industrial scenarios like search and recommendation systems, SRMs facilitate pinpointing which evaluation dimension—whether relevance, timeliness, authority, or diversity—is causing suboptimal performance. Such modular interpretability allows for targeted optimization of dimension-specific components, ensuring adaptability and scalability in single-domain tasks prevalent in industry. Moreover, the framework’s modular design supports parallel computations, significantly improving inference and evaluation efficiency compared to GRMs’ sequential decoding approach.
To validate the practical significance of SRMs, we conducted extensive experiments on public datasets and industrial benchmarks. The results demonstrate that SRMs outperform both scalar RMs and GRMs in accuracy, robustness, and alignment with human preferences. Furthermore, the modular architecture proves highly effective in diagnosing dimensional errors, enabling efficient optimization strategies for real-world applications.

In summary, our contributions are threefold: (1) We systematically analyze the limitations of traditional scalar RMs and GRMs, particularly their inadequate utilization of contextual information and inefficiency in industrial scenarios; (2) We propose the novel Structural Reward Model (SRM), which employs side-branch models as interpretable feature generators to address these challenges; and (3) We validate the effectiveness of SRMs through extensive experiments, showcasing improvements in interpretability, efficiency, and scalability for industry applications.

\section{Related Work}

\textbf{Reward models~(RMs) and Verifiers.} Traditionally, RMs and verifiers are trained as discriminative models through binary classification: given a prompt and a corresponding solution (or a pair of corresponding solutions), the model predicts either the correctness of the solution~\citep{cobbe2021training, lightman2023let, wang2023math, uesato2022solving, liu2023time,wang2022dabert,luo2024improve,liang2019asynchronous,liu2023local, yu2024ovm} or the preference between the two solutions~\citep{stiennon2020learning, nakano2021webgpt}. Concretely, the RM directly produces a numerical continuous-valued score, which is then plugged into a classification objective.

\noindent\textbf{LLM-as-a-Judge.} Verification as next-token prediction by \emph{prompting} off-the-shelf LLMs to serve as a verifier, using either  template\citep{zheng2024judging, bai2022constitutional, song2022improving,kim2023prometheus,gui2018transferring,liang2019adaptive, deductive-verification} or many-shot in-context learning examples~\citep{agarwal2024many}, but \emph{without} specific training for the same~\cite{li2024generation,ma2022searching, gu2024survey,xue2024question, hu2025training,wang2025not}. Our experiments reveal that employing more powerful LLMs as a judge functions worse than trained RM using weaker Gemma models. This finding underscores the critical importance of training verifiers, potentially due to more accurately calibrated uncertainty estimates~\citep{kapoor2024large}.  More broadly, even the strong proprietary LLMs, such as GPT-4~\citep{gpt-4-technical-report} and Gemini ~\citep{team2024gemini}, fall behind trained RMs on popular leaderboards~\citep{rewardbench},  this gap is  larger for reasoning. 

\noindent\textbf{Using CoTs inreward models.} Piror research has explored leveraging CoT reasoning to extract preference and verification signals using LLM-as-a-Judge~\citep{self-reward-lm, meta-rewarding-lm, liu2024resolving,self-taught-evaluator, lee2023rlaif, xue2023dual,sharma2024critical}. Some methods rely on high-quality human data to train critique models~\citep{self-critique-models}, while others focus on training \emph{discriminative} RMs for generating code critiques~\citep{critic-gpt}. For instance, \citet{improve-rm-synthetic-critique} employs CoTs from a separate highly-capable LLM to enhance reward models. The current work of \citet{ankner2024critique} trains an RM to generate response critiques for the generated preference pairs using a more capable LLM. These critiques are then passed as input into a discriminative RM head, separate from the base LLM~\cite{cui2023ultrafeedback, wu2025progressive,dubois2023alpacafarm}. However, these approaches neither unify generation and verification nor filter synthetic critiques for correctness, risking unreliable CoTs in reasoning.

\begin{figure*}[t]
    \centering
    \includegraphics[width=1.0\textwidth]{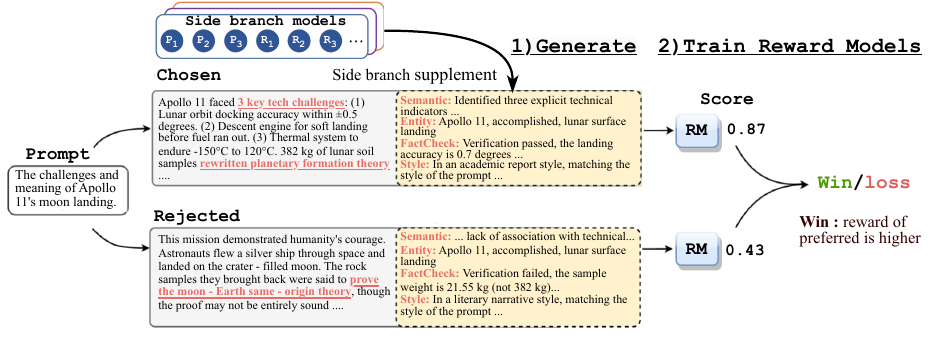} 
    \caption{Overview of the proposed Enhanced Structural Reward Model(SRM) framework integrated with Side Branch Models (SBMs). Given a prompt and candidate responses, the SBMs (Semantic Understanding, Entity Expansion, Fact-Checking, Style Matching, Quality Assessment) generate auxiliary textual contexts. These contexts augment the original input pairs, enabling the Reward Model to deliver evaluations with enhanced accuracy, robustness, and alignment with human preferences.}
    \label{fig:Overview2}
\end{figure*}

\noindent\textbf{Unified Generation and Verification.} DPO and GRPO~\citep{rafailov2024direct,zheng2022robust,guo2025deepseek} and its application to learning verifiers in reasoning~\citep{hosseini2024v} aim to implicitly represent the reward using the logits of a policy that is trainable by reward-modeling loss. However, this approach has been shown to exhibit erroneous extrapolation in learned representations. Prior work has attempted to address these issues with additional techniques, such as iterative reasoning~\citep{iterative-reasoning-preference-op,fei2022cqg,wu2025tablebench}, reinforcement learning on incorrect trajectories~\citep{rl-on-incorrect}, and regularization methods~\citep{pal2024smaug, li2024local,liu2023time,li2024comateformer,yang2024regularizing}. Notably, \citet{yang2024regularizing,wu2025unleashing} train a reward model with an auxiliary generative SFT loss. Unlike approaches that unify generation and verification, these methods avoid text generation during RM queries and rely on more complex training procedures~\cite{meng2024simpo,mao2025mapn}.

\section{Methodology}
To address the limitations of traditional RM in capturing contextual and background information during the inference, we propose an enhanced \textbf{Structural Reward Model (SRM)} framework integrated with Side Branch Models \textbf{(SBMs)}, as illustrated in Figure\ref{fig:Overview2}. The framework leverages SBMs to generate auxiliary features, thereby augmenting the information available to the RM when evaluating the responses. The overall process is as follows: First, sample and filter the training data to obtain high-quality datasets for SBMs training. The SBMs are then applied to the input prompt, chosen response, and rejected response to generate auxiliary features. Finally, these features are concatenated with the original <prompt, chosen> and <prompt, rejected> pairs and fed into the standard RM for classification, as datailed in Algorithm\ref{alg:framework}.






\subsection{Design of Side-Branch Models}

We design five different functional side-branch models, all of which are based on the LLaMA3-8B large language model and obtained through LoRA~\citep{hu2022lora} fine-tuning. The design motivation and details are shown in the Appendix~\ref{sbm}. The specific types of side-branch models are as:
\begin{itemize}[leftmargin=0pt]
\item 1) \textit{Semantic Understanding Model} (SB-Semantic): Extracts the deep semantic information from the <prompt, response> pair,  revealing underlying thematic structures.
\item 2) \textit{Entity Background Information Expansion Model} (SB-Entity): Leverages external knowledge graphs to expand the knowledge background of the core entities and their relational dynamics within the prompt and response.
\item 3) \textit{Fact-Checking Model} (SB-FactCheck): Verifies whether the factual statements in the response are consistent with the known facts and outputs an accuracy analysis text automatically.
\item 4) \textit{Style Matching Analysis Model} (SB-Style): Analyzes the style, tone, and wording of the response, evaluates its uniformity with the style of the prompt, and generates an analysis of the style similarity.
\item 5) \textit{Quality Assessment Model} (SB-Quality): Provides evaluation feedback on the diversity and creativity of the response to avoid generating single and repetitive content.
\end{itemize}

The methodology for collecting and cleaning the training data for side-branch models comprises the following systematic steps:
Initially, we employ the Best-of-N (BoN) sampling strategy on a large-scale prompt-response dataset to generate a comprehensive set of preliminary (prompt, response, auxiliary-text) training candidate triples:
\begin{equation}
    \mathcal{D}_{\text{auxiliary-candidate}} = \{(p,r,a^{(i)}) \mid i = 1,\dots,M\}
\end{equation}
Subsequently, to ensure training data quality, we implement the "LLM-as-a-judge", utilizing a high-performance LLaMa3-8B judge model \footnote{\url{https://platform.openai.com/chat?models=o1}}(denoted as $o_1$) for rigorous quality assessment and screening of candidate data. For each candidate data point ((p, r, $a^{(i)}$)), we input the triple into the judge model to generate a quality score (q) ranging from ([0, 1]):

\begin{equation}
    q = o_1(p, r, a^{(i)}),
\end{equation}
The score magnitude directly correlates with the auxiliary text's ($a^{(i)}$) quality and potential utility for side-branch model training. By establishing a predefined threshold ($\tau$), we selectively retain only auxiliary texts exceeding this quality benchmark:
\begin{equation}
    \mathcal{D}_{\text{auxiliary}} = \{(p, r, a^{(i)}) \mid o_1(p, r, a^{(i)}) \geq \tau\},
\end{equation}

Ultimately, we derive a refined, high-quality side-branch training dataset ($\mathcal{D}_{\text{auxiliary}}$), which serves as the foundation for fine-tuning the corresponding side-branch model ($SB_i$) through maximum likelihood optimization:


\begin{equation}
    \begin{aligned}
        \mathcal{L}_{\text{SB}_i}(\phi_i) &= -\frac{1}{|\mathcal{D}_{\text{auxiliary}}|} \\
        &\quad \sum_{(p,r,a)\in\mathcal{D}_{\text{auxiliary}}}\log P_{\phi_i}(a\mid p,r),
    \end{aligned}
\end{equation}
where $\phi_i$ denotes the parameters of the $i$-th side-branch model, and $a$ represents the generated auxiliary text based on the prompt-response pair $(p,r)$.


\subsection{Construction and Training of the Enhanced Reward Model}

Following the training of each side-branch model, we concatenate the output texts of the side-branch models with the original text pairs <prompt, chosen> and <prompt, reject> and , thereby obtaining the enhanced input representations:


\begin{equation}
\begin{aligned}
    x_{\text{chosen}}=p\oplus r_c\oplus t_c^{(1)}\oplus t_c^{(2)}\dots\oplus t_c^{(N)},\\
    x_{\text{reject}}=p\oplus r_j\oplus t_j^{(1)}\oplus t_j^{(2)}\dots\oplus t_j^{(N)},
    \end{aligned}
\end{equation}

where $t_c^{(i)}$ and $t_j^{(i)}$ represent the auxiliary texts generated by the $i$-th side-branch model with ($p, r_c$) and ($p, r_j$) as inputs respectively.

The standard Reward Model (RM) takes the enhanced inputs (i.e., $x_{\text{chosen}}$ and $x_{\text{reject}}$) and then calculates the score values respectively:
\begin{equation}
    s_c=RM(x_{\text{chosen}};\theta), \quad s_j=RM(x_{\text{reject}};\theta),
\end{equation}

\begin{table*}
    \centering
    \small
    \renewcommand{\arraystretch}{1.2}
\setlength{\tabcolsep}{8pt}
    \begin{tabular}{lccccccc}
    \toprule
    \multirow{2}{*}{Model} & \multicolumn{2}{c}{RM-Bench} & \multirow{2}{*}{JudgeBench} & \multicolumn{3}{c}{IFBench} & \multirow{2}{*}{Overall} \\
    \cmidrule{2-3} \cmidrule{5-7}
    & Normal & Hard & & Simple & Normal & Hard & \\
    \midrule
ArmoRM-Llama3-8B-v0.1 &$76.7$&$34.6$&$51.9$&$72.3$&$66.2$&$59.5$&$56.5$\\
INF-ORM-Llama3.1-70B &$77.5$&$25.1$&$59.1$&$78.7$&$69.2$&$53.8$&$55.7$\\
Skywork-Reward-Llama-3.1-8B-v0.2 &$78.0$&$31.8$&$57.8$&$78.7$&$69.2$&$59.8$&$58.1$\\
Skywork-Reward-Gemma-2-27B &$82.7$&$35.1$&$55.8$&$87.2$&$68.4$&$56.1$&$59.2$\\
Openai-GPT-4o &$71.4$&$27.9$&$64.6$&$85.1$&$66.2$&$54.4$&$56.3$\\
Openai-GPT-4o mini &$60.5$&$15.0$&$51.9$&$70.2$&$59.4$&$51.9$&$45.9$\\
\midrule
Llama3-8B Instruct 
&$\phantom{0}9.3$&$20.2$&$\phantom{0}2.6$&$12.8$&$12.8$&$13.6$&$11.3$\\
\quad w/ side-branch (SRM) &$75.4$&$39.5$&$59.4$&$77.1$&$63.6$&$56.1$&$60.8$\\
\midrule
internlm2-7b-reward &$72.6$&$19.9$&$56.2$&$74.5$&$61.7$&$55.7$&$52.0$\\
\quad w/ side-branch (SRM) &$78.4$&$46.8$&$58.7$&$75.1$&$66.9$&$62.2$&$63.1$\\
internlm2-20b-reward &$74.4$&$26.1$&$61.7$&$74.5$&$68.4$&$58.7$&$56.4$\\
\quad w/ side-branch (SRM) &$79.1$&$47.4$&$59.8$&$76.5$&$68.7$&$64.6$&$64.3$\\
    \bottomrule
    \end{tabular}
    \caption{
    The experimental results (\%) of all investigated baselines and our proposed method. The overall score is calculated as the average of RM - Bench, JudgeBench, and the averaged score across three subsets of IFBench.}
    \label{tab:main_exp}
\end{table*}


where $\theta$ represents the trainable parameters of the reward model. These scores, $s_c$ and $s_j$, quantify the model's preference for the chosen and rejected responses, respectively.

The optimization objective of the Bradley-Terry reward model is grounded in the Bradley-Terry pairwise comparison framework, which models the probability of the chosen response being preferred over the rejected response:
\begin{equation}
    \mathcal{L}_{\text{BT-RM}}(\theta)=-\frac{1}{|\mathcal{D}_{\text{t}}|}\sum_{(p,r_c,r_j)\in\mathcal{D}_{\text{t}}}\log P(r_c \succ r_j | p),
\end{equation}
where the probability $P(r_c \succ r_j | p)$ is defined as:
\begin{equation}
    P(r_c \succ r_j | p) = \frac{e^{s_c}}{e^{s_c} + e^{s_j}}.
\end{equation}
By minimizing this loss, the Bradley-Terry Reward Model learns to capture human preferences and generates more accurate evaluations.

\section{Experiments}

\subsection{Experimental Setup}
\label{sec:exp_setup}

\paragraph{Evaluation Benchmarks} Reward model benchmarks typically comprise an instruction-response pair, with the primary objective of identifying the superior response. We evaluate our approach on RM-Bench~\citep{liu2024rm}, JudgeBench~\citep{tan2024judgebench}, and a novel benchmark \ourdataset. RM-Bench and JudgeBench include response pairs that evaluate factual accuracy, with the former's chat subset used under both standard and challenging settings, and the latter's knowledge subset prioritized. \ourdataset is designed to assess how well RMs prioritize instruction-constrained response, aligning with the framework in \cite{peng2025agentic}.

\paragraph{Baselines}
\looseness = -1

We compare our approach against two baseline categories: (1) Regression-based RMs: Specifically trained to score responses and select the highest-ranked candidates, including advanced models such as ArmoRM~\citep{wang2024interpretable}, INF-ORM-Llama3.1-70B~\citep{infly2024inf}, Skywork-Reward~\citep{liu2024skywork}, and internlm2 reward~\citep{cai2024internlm2}. (2) Generative LLM-based RMs: Leveraging Large Language Models for response scoring or pairwise comparisons performing to identify the best response~\citep{lambert2024rewardbench}. We evaluate across proprietary models like GPT-4o~\citep{OpenAI20244o} and GPT-4o mini~\citep{OpenAI2024}, as well as open-source variants such as Llama3-8B-Instruct~\citep{dubey2024llama}. And for detailed comparison results and computational efficiency comparison with GRM, see Appendix~\ref{sbm2}.

\subsection{Experimental Results}
\label{sec:exp_result}
The experimental results from Table~\ref{tab:main_exp} demonstrates that Structural Reward Model substantially improves performance across benchmarks. 
First, the overall performance shows that introducing the side branch notably boosts the scores across all base models. For instance, the Llama3-8B Instruct model's overall score increases sharply from 11.3\% to 60.8\%. Similarly, the Internlm2-7B-Reward and Internlm2-20B-Reward models achieve significant gains of 11.1\% and 7.9\%, respectively, after applying our method. 
Second, in the RM-Bench benchmark, our side branch consistently delivers substantial performance improvements under both Normal and Hard settings. Specifically, under the Normal difficulty, the score of Llama3-8B Instruct rises from 9.3\% to 75.4\%, while it improves from 20.2\% to 39.5\% under the more challenging Hard level. This trend persists for Internlm2-based models, with Internlm2-20B-Reward showing a substantial increase from 26.1\% to 47.4\%, especially on the Hard setting. 
Third, in the JudgeBench knowledge subset evaluation, the side branch method provides consistent and positive gains. For example, Llama3-8B-Instruct improves from an initial 2.6\% to 59.4\%. Similarly, Internlm2-7B-Reward shows an improvement of 2.5\%, and although Internlm2-20B-Reward exhibits a slight decrease of 1.9\%, it still maintains a relatively high overall performance. 
Finally, on the newly proposed IFBench benchmark across three different subsets (Simple, Normal, and Hard), adding the side branch clearly enhances performance, particularly in the more challenging Normal and Hard subsets. For instance, Internlm2-7B-Reward achieves increases of 5.2\% on the Normal level and 6.5\% on the Hard level, while Internlm2-20B-Reward gains an evident improvement of 5.9\% in the Hard subset.

\subsection{Ablation Study}
\label{subsec:ablation}

\begin{figure}[t]
    \centering
    \includegraphics[width=0.495\textwidth]{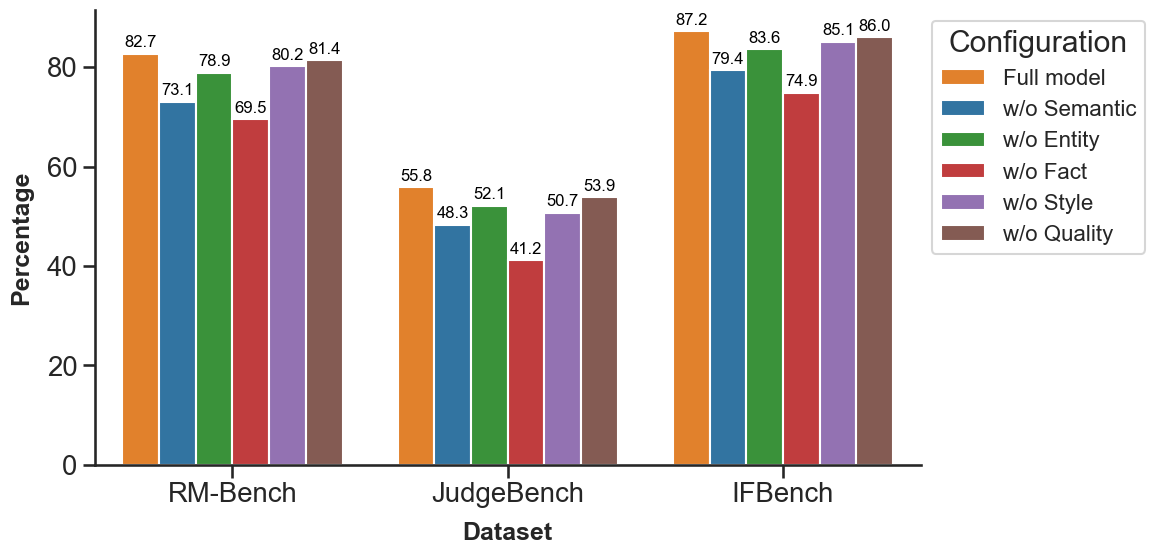} 
    \caption{Ablation study of side branch models (\%)}
    \label{fig:Figure3}
\end{figure}

To evaluate the individual contributions of each side branch module within the enhanced structural reward model (SRM), we conducted an ablation study across three benchmarks: RM-Bench, JudgeBench, and IFBench, as summarized in Table~\ref{fig:Figure3}. The removal of the \textbf{Fact-Checking} module precipitated the most substantial performance declines of 13.2\%, 14.6\%, and 12.3\%, respectively. This underscores its critical role in ensuring factual consistency, which directly influences the RM's ability to discriminate between correct and incorrect responses. The Semantic Understanding module also proved pivotal, with its exclusion causing significant performance losses (9.6\%, 7.5\%, and 7.8\%), confirming its essential function in alignment responses with the prompt's context and intent.
Moreover, removing auxiliary modules such as Entity Expansion, Style Matching, and Quality Assessment resulted in smaller yet discernible performance declines (ranging from 1.3\% to 5.1\%), indicating their supportive roles in capturing nuanced response characteristics like richness, stylistic appropriateness, and clarity. Interestingly, the impact of module removal varied across benchmarks. JudgeBench demonstrated heightened sensitivity to Fact-Checking module removal, reflecting its emphasis on factual correctness. Conversely, RM-Bench and IFBench exhibited greater reliance on Semantic Understanding, aligning with their focus on contextual and comprehensive evaluation.
These findings collectively validate the modular design of our framework, where core modules like Fact-Checking and Semantic Understanding constitute the backbone of the RM's performance, while auxiliary modules refine the evaluation by addressing complementary quality dimensions.


\subsection{Case Study}
\label{subsec:CaseStudy}
To demonstrate the effectiveness of integrating SBMs into the SRM, we present a qualitative case study in Table \ref{tab:case_study}. 
The baseline reward model (RM) without SBMs erroneously favors the rejected response, assigning it a higher score (0.68 vs. 0.52) due to inadequate semantic and factual understanding. This underscores the limitations of conventional RMs, which rely solely on surface-level textual features while failing to incorporate contextual information.
However, after integrating the proposed side branch models into the reward model, we achieve significant improvements in evaluation accuracy. Specifically, the SBM modules individually provide critical contextual insights: \textit{Semantic Understanding} identifies temporal relevance and conceptual alignment with the prompt;  \textit{Entity Expansion} provides additional entity-level information (e.g., coffee's cardiovascular benefits).
These enhancements enable the SBM-augmented RM to prioritize the chosen response with markedly higher accuracy (0.91 vs. 0.32), objectively reflecting factual correctness, updated evidence, and semantic coherence. The scoring adjustment validates the effectiveness of our SBM-enhanced methodology.


\section{Conclusion}

In this paper, we introduced the \textbf{Structural Reward Model (SRM)}, a novel approach to address the limitations of traditional scalar RMs and Generative RMs (GRMs) in reward modeling tasks. Unlike scalar RMs, which rely solely on prompt-response pairs, and GRMs, which operate as black-box generators, SRMs leverage modular and interpretable side-branch models to generate auxiliary features that capture fine-grained contextual signals. This structured and modular design enables SRMs to provide domain-specific, dimension-aware evaluations, making them particularly suitable for industrial scenarios such as search and recommendation systems. 
Extensive experiments conducted on public datasets and industrial benchmarks validate the practical significance of SRMs, demonstrating superior performance in accuracy, robustness, alignment with human preferences, and dimensional error diagnosis compared to both scalar RMs and GRMs.
And we hope SRM can inspire further innovation in structured, modular approaches.
\newpage

\section{Limitations}
Despite the strong performance of our proposed SRM on multiple benchmarks, the framework has several limitations. The reliance on a set of predefined Side-Branch Models (SBMs) tailored for specific dimensions means their design requires significant domain knowledge, extensive tuning, and more computational resources than a  scalar reward model. Additionally, the framework's effectiveness is highly dependent on high-quality training data, and our employment of  "LLM-as-a-judge" strategy  to score and filter data could introduce potential noise or bias. Finally, the current feature fusion method of concatenating auxiliary texts with the original input may not be optimal, as it can create excessively long sequences and increase the model's processing load, suggesting that more efficient fusion mechanisms could be explored in the future.

\bibliography{custom}

\begin{table*}[ht]
\centering
\small
\renewcommand{\arraystretch}{1.6}
\setlength{\tabcolsep}{4pt} 

\definecolor{headercolor}{HTML}{D9EAF7} 
\definecolor{rowodd}{HTML}{F3F7FA}      
\definecolor{highlight}{HTML}{DEF7E8}  

\begin{tabular}{lcccccc}
\toprule
\rowcolor{headercolor} 
\textbf{RL Method} & \textbf{RM Type} & \textbf{Accuracy (\%)} & \textbf{Knowledge (\%)} & \textbf{Hallucination (\%)}$\downarrow$ & \textbf{Creativity (\%)} & \textbf{Complex (\%)} \\
\midrule

\rowcolor{rowodd} 
\multirow{2}{*}{\textbf{DPO}} 
& Vanilla-RM & 78.1 & 78.8 & 14.5 & 75.2 & 61.2 \\
& \cellcolor{highlight}\textbf{SB-RM (ours)} & \textbf{81.6} (\textcolor{blue}{+3.5}) & \textbf{81.9} (\textcolor{blue}{+3.1}) & \textbf{8.6} (\textcolor{blue}{$-$5.9}) & \textbf{79.5} (\textcolor{blue}{+4.3}) & \textbf{67.5} (\textcolor{blue}{+6.3}) \\

\rowcolor{white}
\midrule

\rowcolor{rowodd} 
\multirow{2}{*}{\textbf{PPO}} 
& Vanilla-RM & 81.7 & 80.6 & 15.1 & 77.8 & 62.7 \\
& \cellcolor{highlight}\textbf{SB-RM (ours)} & \textbf{84.0} (\textcolor{blue}{+2.3}) & \textbf{82.1} (\textcolor{blue}{+1.5}) & \textbf{8.8} (\textcolor{blue}{$-$6.3}) & \textbf{81.7} (\textcolor{blue}{+3.9}) & \textbf{68.9} (\textcolor{blue}{+6.2}) \\

\rowcolor{white}
\midrule

\rowcolor{rowodd} 
\multirow{2}{*}{\textbf{GRPO}} 
& Vanilla-RM & 82.2 & 80.6 & 13.7 & 78.5 & 62.9 \\
& \cellcolor{highlight}\textbf{SB-RM (ours)} & \textbf{84.4} (\textcolor{blue}{+2.2}) & \textbf{82.4} (\textcolor{blue}{+1.8}) & \textbf{8.2} (\textcolor{blue}{$-$5.5}) & \textbf{82.8} (\textcolor{blue}{+4.3}) & \textbf{68.4} (\textcolor{blue}{+5.5}) \\

\bottomrule
\end{tabular}

\caption{\small
Comparative Analysis of Vanilla Reward Model and Structural Reward Model in Industrial Settings. The evaluation compares the performance of the Vanilla Reward Model (Vanilla-RM) and our proposed Side-Branch Enhanced Reward Model (SRM) using a comprehensive black-box test set of 150,000 realistic industrial samples.
Key performance dimensions are highlighted with a green background, including: - Accuracy - Factual Knowledge - Hallucination Reduction - Creativity - Complex Reasoning.
\textcolor{blue}{Blue values} indicate the absolute percentage point improvements relative to the baseline model.}
\label{tab:real-world-results-enhanced}
\end{table*}

\begin{table*}[h]
\centering
\renewcommand{\arraystretch}{1.52}
\setlength{\tabcolsep}{10pt}

\definecolor{headerbg}{RGB}{230, 245, 255}
\definecolor{rowbg}{RGB}{245, 249, 255}
\definecolor{textcolor}{RGB}{50, 90, 150}

\begin{tabular}{>{\columncolor{headerbg}}p{4cm}|>{\columncolor{rowbg}}p{5cm}|>{\columncolor{rowbg}}p{5cm}}
\toprule
\rowcolor{headerbg}\textbf{\textcolor{textcolor}{Category}}                       & \multicolumn{1}{c|}{\textbf{\textcolor{textcolor}{Original RM (Without SBMs)}}} & \multicolumn{1}{c}{\textbf{\textcolor{textcolor}{Structura RM (With SBMs)}}} \\ 
\midrule
\rowcolor{rowbg}\textbf{\textcolor{textcolor}{Prompt (p)}}                     & \multicolumn{2}{p{10cm}}{\textit{"Discuss the health effects of daily caffeine consumption."}} \\ 

\rowcolor{white}\textbf{\textcolor{textcolor}{Chosen Response (r\_c)}}         & \multicolumn{2}{p{10cm}}{\textit{"Moderate caffeine intake (300-400mg/day) may enhance cognitive performance. Recent studies suggest potential cardiovascular benefits when consumed without added sugars (NIH, 2023)."}} \\ 

\rowcolor{rowbg}\textbf{\textcolor{textcolor}{Rejected Response (r\_j)}}       & \multicolumn{2}{p{10cm}}{\textit{"Coffee causes heart disease and bone loss. A 1995 study proved caffeine directly weakens bones (Journal of Old Medicine)."}} \\ 
\midrule
\rowcolor{white}\textbf{\textcolor{textcolor}{Semantic Understanding}}         & Detected mismatch: outdated "1995 study," there is a timeliness issue.   & Modern research shows that the semantic correlation between response and prompt is high. \\ 

\rowcolor{rowbg}\textbf{\textcolor{textcolor}{Entity Expansion}}               & < Coffee; Disadvantages; Stimulates the stomach and intestines and affects the digestive system >                 & < Coffee; Benefits; Reduced Risk of Cardiovascular Disease > \\

\rowcolor{white}\textbf{\textcolor{textcolor}{Fact Checking}}                  & Verification failed, Flagged retracted study: “J OldMed (1995) retracted in 2005"              & Verification passed, the content is factually correct. \\

\rowcolor{rowbg}\textbf{\textcolor{textcolor}{Style Analysis}}                 & Single style-Academic style    & Single style-Academic style  \\

\rowcolor{white}\textbf{\textcolor{textcolor}{Quality Assessment}}             & There is no repetition or redundant expression, and the key information can be conveyed efficiently.                     & There is no repetition or redundant expression, and the key information can be conveyed efficiently. \\ 
\midrule
\rowcolor{rowbg}\textbf{\textcolor{textcolor}{Reward Model Scores}}            & \multicolumn{1}{c|}{\textbf{\textcolor{textcolor}{r\_c: 0.52 \quad r\_j: 0.68}}}   & \multicolumn{1}{c}{\textbf{\textcolor{textcolor}{r\_c: 0.91 \quad r\_j: 0.32}}} \\ 

\rowcolor{white}\textbf{\textcolor{textcolor}{Final Judgment}}                 & \multicolumn{1}{c|}{\textbf{\textcolor{textcolor}{Incorrect: Preferred r\_j}}} & \multicolumn{1}{c}{\textbf{\textcolor{textcolor}{Correct: Preferred r\_c}}} \\ 
\bottomrule
\end{tabular}
\caption{Performance Evaluation: Reward Model Enhancement through Side-Branch Model Integration.
Comparative analysis of reward model performance before and after integrating Side-Branch Models (SBMs), assessing: - Semantic Understanding - Entity Expansion - Fact-Checking - Stylistic Alignment - Overall Response Quality.
SBM integration significantly improves reward model accuracy in discriminating response quality.
}
\label{tab:case_study}
\end{table*}

\newpage

\appendix


\section{Evaluation in Industrial Applications}

We conducted extensive evaluations of our proposed Side-Branch  Models enhanced  Structural Reward Model (SRM) within a realistic industrial scenario to assess its effectiveness in practical deployment environments.

\subsection{Experiment Settings}

\paragraph{Industrial Training Dataset.} We constructed a real-world dataset comprising 1.8 million preference-labeled samples covering diverse scenarios frequently encountered in practical deployments, including mathematics, code generation, reasoning, instruction-following, STEM domains, standard NLP tasks, factual knowledge verification, hallucination control, multilingual applications, creative generation, and professional domain tasks. The dataset was meticulously annotated and reviewed by professional human annotators to ensure practical relevance and annotation accuracy.

 \paragraph{Industrial Evaluation Dataset.} Models were evaluated on an independent black-box test set consisting of approximately 150,000 annotated samples, representing the same usage scenarios as the training dataset but strictly excluded from the training process.

\paragraph{Overall Training Setting.} To ensure robust evaluation, we employed three representative reinforcement learning methods which are widely used in industrial practice: Direct Preference Optimization (DPO), Proximal Policy Optimization (PPO), and Generalized Reinforcement from Preference Optimization (GRPO). For fair comparisons, we trained the same base model (InternLM2-20B) utilizing both our proposed SBM-RM and a vanilla Reward Model (Vanilla-RM).

\section{Results Analysis}
\label{subsec:industrial_results}

Table~\ref{tab:real-world-results-enhanced} summarizes the evaluation results from an industrial deployment scenario. The assessment evaluates critical real-world application metrics, including overall accuracy, factual knowledge precision, hallucination reduction, creativity, and complex reasoning capabilities. We compare our proposed Side-Branch enhanced Structural Reward Model (\textbf{SRM}) against the standard Vanilla Reward Model (\textbf{Vanilla-RM}), employing three representative reinforcement learning algorithms prevalent in industrial practice: Direct Preference Optimization, Proximal Policy Optimization, and Generalized Reinforcement from Preference Optimization.
The results consistently demonstrate that the proposed side-branch integration outperforms the baseline across all metrics and reinforcement learning methods. This confirms the effectiveness and generalizability of diverse auxiliary contexts provided by side branch models in industrial-scale scenarios, and the detailed analysis is as follows:


\paragraph{Accuracy and Knowledge Enhancement.} Compared to baseline reward models, side branches notably enhance overall response accuracy and factual knowledge correctness. Under DPO training, accuracy increases by \textcolor{blue}{3.5\%} and factual knowledge precision improves by \textcolor{blue}{3.1\%}. PPO and GRPO training methods similarly show clear improvements, validating the robust contributions of side branch models (SB-Semantic and SB-Entity) in providing enriched semantic and contextual information.

\paragraph{Hallucination Mitigation.} A critical challenge in industrial LLM deployments is model hallucination. Our SRM significantly and consistently reduces hallucination rates across experimental settings, with decreases of \textcolor{blue}{5.9\%} under DPO, \textcolor{blue}{6.3\%} under PPO, and \textcolor{blue}{5.5\%} under GRPO. This validates the SB-FactCheck side branch's effectiveness in penalizing hallucinations and producing more accurate and trustworthy model outputs.

\paragraph{Improvements in Creativity and Complex Reasoning.} Our framework demonstrates clear improvements in creativity and complex reasoning benchmarks. Systematic gains are observed across all evaluated reinforcement learning methods, with creativity measures increasing up to \textcolor{blue}{4.3\%} under DPO and GRPO, and complex reasoning performances increasing approximately \textcolor{blue}{6.3\%(DPO)}, \textcolor{blue}{6.2\%(PPO)}, and \textcolor{blue}{5.5\%(GRPO)}. These improvements highlight the practical utility of SB-Quality and SB-Semantic models in assessing diverse, innovative, and reasoning-intensive model outputs.


In summary, the consistent superior performance across multiple benchmarks establishes the practical effectiveness of our enhanced Reward Model framework integrating Side Branch Models. The robustness across diverse reinforcement learning settings indicates significant generalizability and practical merits for industrial deployments.

\begin{table*}[htbp]
\centering
\small
\renewcommand{\arraystretch}{1.6}
\setlength{\tabcolsep}{14pt}
\definecolor{headercolor}{HTML}{D9EAF7}
\definecolor{rowodd}{HTML}{F3F7FA}
\caption{Industrial Defect Patterns and Corresponding Models}
\label{tab:defect_mapping}
\begin{tabular}{lccc}
\toprule
\rowcolor{headercolor}
\textbf{Model} & \textbf{Core Issue} & \textbf{Representative Case} & \textbf{Frequency} \\
\midrule
\rowcolor{rowodd}
SB-Semantic & Semantic mismatch & "Portable charger" vs "Power bank" mismatch & 38.7\% \\
\rowcolor{rowodd}
SB-Entity & Knowledge deficiency & Missing graphene fabric properties & 22.1\% \\
\rowcolor{rowodd}
SB-FactCheck & Factual inconsistency & Overstated battery life claims & 15.4\% \\
\rowcolor{rowodd}
SB-Style & Tone discordance & Technical specs in casual language & 9.8\% \\
\rowcolor{rowodd}
SB-Quality & Content redundancy & Repeated similar recommendations & 13.2\% \\
\bottomrule
\end{tabular}
\end{table*}

\section{Side-Branch Model Design Rationale} 
\label{sbm}

\subsection{Industrial-Driven Design Methodology}
When designing and optimizing Side-Branch Models (SBMs), we follow an industry-inspired, case-driven iterative process. Specifically, we conduct $N$-fold cross-validation on the training data to expose and analyze "bad cases," and then perform attribution analysis to identify core problem categories. These are further abstracted into generalizable issue patterns, which directly inform the targeted design of SBMs. This approach achieves a balance between addressing practical requirements and providing theoretical support, ensuring that the models not only tackle existing issues but also maintain generality and extensibility.
Through extensive investigation on public training and evaluation datasets, we observe several recurring challenges that frequently cause discrepancies between reward model evaluation and human expectations:1) Difficulty in capturing deep semantic meaning or nuanced topics, leaving latent user intents unsatisfied; 2)Insufficient understanding of entity and relational background knowledge, compromising assessment of relevance; 3)Lack of fact consistency, leading to undetected factual errors or questionable statements; 4)Style or tone mismatches between response and prompt, degrading user experience; 5)Repetitive or low-diversity, lacking in novelty and reducing user satisfaction.

\subsection{Defect-Centric Model Construction}
Table~\ref{tab:defect_mapping} demonstrates how each side-branch model corresponds to specific industrial pain points. The SB-Semantic model addresses the most frequent issue (38.7\% occurrence) where literal keyword matching fails to capture query intent, such as mismatching "waterproof sports earphones" with non-waterproof products. Our solution employs domain-adapted semantic encoding through a fine-tuned sentence-BERT model, calculating semantic similarity via $f_\theta(p,r) = \text{Sim}_{\text{cosine}}(E_d(p), E_d(r))$ where $E_d$ represents our proprietary encoder.
The SB-Entity branch combats knowledge gaps in 22.1\% of cases through real-time knowledge graph augmentation. For each entity $e$ in prompt/response pairs, we retrieve contextual knowledge $\mathcal{K}(e) = \bigcup \text{Neighbor}(e) \cup \text{Attributes}(e)$ from a dynamically updated product KG that synchronizes with new item listings hourly.

\subsection{Detail of the Five SBM Types}

\textbf{SB-Semantic (Semantic Understanding Model):} The motivation arises from numerous real-world cases where reward models tend to perform superficial pattern matching instead of deep semantic comprehension, making them insensitive to subtle distinctions or implicit intents within responses. Attribution stems from both manual quality inspection results and user feedback, highlighting that deeper semantic understanding is crucial for raising relevance and user satisfaction.
\textbf{SB-Entity (Entity Background Enrichment Model):} In practice, many tasks involve domain-specific background, proper nouns, or specialized entities. Deficiencies in external knowledge frequently render the reward model unable to deliver accurate judgments. Attribution analysis in high-background domains (e.g., finance, medicine, law) reveals that introducing knowledge graph or external knowledge enrichment can significantly mitigate such weaknesses.
\textbf{SB-FactCheck (Fact-Checking Model):} For domains like medical QA, news dialogue, or professional counseling, factual errors may result in severe practical consequences (e.g., misleading users). Online negative feedback and user reports show that most "critical" bad cases involve factual inconsistency, necessitating explicit fact-checking side-branch intervention.
\textbf{SB-Style (Style Matching Analysis Model):} Users often have clear preferences regarding interaction style, tone, and degree of professionalism. If the reward model evaluates only content while ignoring appropriate style matching, user discomfort or a decline in perceived trustworthiness may ensue. For customer service and omni-channel dialog scenarios, frequent negative feedback can be traced to mismatched or inappropriate response style.
\textbf{SB-Quality (Quality Assessment Model):} Repetitive, low-quality, or unoriginal answers have a markedly negative impact on user engagement and platform reputation. Bad case analysis demonstrates that substantial user complaints stem from lack of diversity or novelty, underscoring the necessity of a dedicated assessment branch for these aspects.

\begin{algorithm}[ht]
\setstretch{0.75}      
\SetNlSkip{-0.3em}     
\DontPrintSemicolon   

\Input{Prompt$p$,Chosen Response $r_c$,Rejected Response $r_j$, Side-Branch Models $SB_i$, Judge Model $o_1$, Threshold $\tau$}
\Output{Enhanced Reward Model $RM$}

\vspace{0.1em}
\textbf{Step 1: Train Side-Branch Models}
\begin{itemize}[leftmargin=*,nosep] 
    \item Sample candidate data:
    \vspace{-0.5em} 
    \begin{equation*}
    \mathcal{D}_{\text{auxiliary-candidate}} = \{(p, r, a^{(i)}) \mid i = 1,\dots,M\}
    \end{equation*}
    
    \item Filter using $o_1$:
    \vspace{-0.5em}
    \begin{equation*}
    \mathcal{D}_{\text{auxiliary}} = \{(p, r, a^{(i)}) \mid o_1(p, r, a^{(i)}) \geq \tau\}
    \end{equation*}
    
    \item Fine-tune $SB_i$:
    \vspace{-0.5em}
    \begin{equation*}
    \mathcal{L}_{\text{SB}_i}(\phi_i) = -\frac{1}{|\mathcal{D}_{\text{a}}|}
    \sum_{\mathclap{(p, r, a) \in \mathcal{D}_{\text{a}}}}
    \log P_{\phi_i}(a \mid p, r)
    \end{equation*}
\end{itemize}

\vspace{0.5em}
\textbf{Step 2: Generate Auxiliary Features}
\begin{itemize}[leftmargin=*,nosep]
    \item Generate texts via $SB_i$:
    \vspace{-0.5em}
    \begin{align*}
    x_{\text{chosen}} &= p \oplus r_c \oplus t_c^{(1)} \oplus \dots \oplus t_c^{(N)} \\
    x_{\text{reject}} &= p \oplus r_j \oplus t_j^{(1)} \oplus \dots \oplus t_j^{(N)}
    \end{align*}
\end{itemize}

\vspace{0.5em}
\textbf{Step 3: Train Enhanced Reward Model}
\begin{itemize}[leftmargin=*,nosep]
    \item Compute scores:
    \vspace{-0.5em}
    \begin{equation*}
    s_c = RM(x_{\text{chosen}}; \theta),\quad 
    s_j = RM(x_{\text{reject}}; \theta)
    \end{equation*}
    
    \item Optimize with loss:
    \vspace{-0.5em}
    \begin{equation*}
    \mathcal{L}_{\text{RM}}(\theta) = -\frac{1}{|\mathcal{D}_{\text{t}}|}\sum_{(p,r_c,r_j)\in\mathcal{D}_{\text{t}}}\log P(r_c \succ r_j | p)
    \end{equation*}
\end{itemize}

\vspace{0.1em}
\Return{RM with SBs enhanced prediction}

\caption{Enhanced Reward Model with Side-Branch Models}
\label{alg:framework}
\end{algorithm}


\begin{table}[t]
\centering
\small
\renewcommand{\arraystretch}{1.6}
\setlength{\tabcolsep}{8pt}
\definecolor{headercolor}{HTML}{D9EAF7}
\definecolor{rowodd}{HTML}{F3F7FA}
\begin{tabular}{lcc}
\toprule
\rowcolor{headercolor}
\textbf{Method} & \textbf{Public Dataset} & \textbf{Industrial Dataset} \\
\midrule
\rowcolor{rowodd}
Scalar RM       & 18.7          & 21.3              \\
\rowcolor{rowodd}
GRM             & 92.5          & 106.1             \\
\rowcolor{rowodd}
SRM (Ours)      & \textbf{22.8} & \textbf{25.4}     \\
\bottomrule
\end{tabular}
\caption{Inference time (seconds per 1,000 samples) for different reward modeling methods.}
\label{tab:efficiency}
\end{table}

\section{Efficiency Improvement}
\label{sbm2}

Efficiency is a critical requirement for reward modeling in industrial applications, where large-scale inference and real-time feedback are essential. Traditional scalar RMs are computationally efficient due to their simple architecture, but often at the cost of limited contextual comprehension. Conversely, GRMs introduce intermediate reasoning but suffer from high computational overhead, primarily because of the sequential decoding inherent to autoregressive generation. This bottleneck is particularly pronounced when evaluating large candidate pools or deploying on latency-sensitive tasks.

The proposed Structural Reward Model (SRM) framework addresses this challenge through its modular and parallelizable design. Unlike GRMs, where every evaluation must generate full reasoning chains before a scalar decision, SRM leverages specialized side-branch models to independently extract auxiliary features from the (prompt, response) pair and related context. Each side-branch model operates as a lightweight, targeted feature extractor, allowing all branches and the main RM to be computed in parallel. This results in significantly reduced inference latency and improved throughput.
Specifically, given $K$ auxiliary dimensions, the SRM initializes $K$ side-branch modules. These modules are fine-tuned for their respective tasks and are optimized for efficient inference. During the evaluation phase, all side-branches simultaneously generate corresponding feature representations, which are then aggregated by a lightweight main RM head to produce the final reward score.

To empirically evaluate the efficiency gains, we benchmark SRM, scalar RM, and GRM on both public benchmarks and proprietary industrial datasets. Table~\ref{tab:efficiency} presents the inference time per 1,000 examples for each reward modeling method. The results indicate that SRM achieves up to \textbf{4$\times$ faster} inference than GRM while providing substantially richer signal for downstream tasks. Moreover, SRM's design enables distributed deployment and scaling, making it suitable for large-scale, high-availability industry environments.
In summary, SRM balances interpretability and contextual awareness with practical efficiency, enabling high-throughput, low-latency inference that is crucial for industrial-scale language model deployment.

\end{document}